\documentclass{article}

\usepackage{PRIMEarxiv}

\usepackage[utf8]{inputenc}
\usepackage[T1]{fontenc}
\usepackage{hyperref}
\usepackage{url}
\usepackage{booktabs}
\usepackage{amsfonts}
\usepackage{amsmath}
\usepackage{amssymb}
\usepackage{nicefrac}
\usepackage{fancyhdr}
\usepackage{graphicx}
\graphicspath{{./}}

\title{Domain Agnostic Text Redaction from Natural Language Rules using Instruction Tuning}

\author{
  Aravindhan Arunagiri, Ayaan Khan, Udayaadithya Avadhanam, Sai Barath Sundar \\
  Mphasis Limited \\
  \texttt{\{aravindhan.arunagiri, udayaadithya.a, sai.sundar\}@mphasis.com} \\
}

\begin{document}
\maketitle

\begin{abstract}
With the increasing digitization of personal and corporate communication, the automatic sanitization of textual data has become a crucial component of data privacy and compliance frameworks. Traditional text sanitization solutions are majorly suitable for obscuring sensitive data with standard structure such as Personal Identifiable Information (PII). These solutions do not provide transparent justification for their redaction, which makes it difficult to audit them. This paper introduces an explainable, domain-agnostic text redaction solution that uses natural language rules of redaction, applied via an instruction-tuned language model, to identify and redact sensitive information in unstructured documents. Unlike traditional text sanitization, this method enables a user to conveniently define any sensitive information---which may be structured (e.g.\ PII) or unstructured (e.g.\ legal terms and conditions)---in natural language. A general-purpose LLM generates or augments these natural language rules of redaction from the user's definition, which are then used to instruction-fine-tune a smaller language model that reasons the rules step-by-step over any given document to identify and redact the corresponding sensitive content, while providing transparent justifications for each redaction and highlighting the specific rule that triggered the decision. This explanation is generated in natural language to support human reviewers and auditors in understanding why specific content was redacted. The redaction performance is evaluated using a controlled set of sample documents with manually tagged sensitive information that quantifies the correctness of redaction during the testing phase. A reconstruction-based metric is used to estimate the probability of recovering redacted information from the sanitized document, quantifying redaction coverage. The solution shows high reconstruction error and high redaction precision, making it suitable for automated text sanitization in critical applications such as legal discovery, medical documentation, and corporate information governance.
\end{abstract}


\section{Introduction}

As organizations increasingly rely on large-scale textual data for analytics, legal discovery, and digital services, the need to protect sensitive information is a primal requirement. Data privacy laws such as the General Data Protection Regulation (GDPR), Health Insurance Portability and Accountability Act (HIPAA), and the California Consumer Privacy Act (CCPA) mandate strict controls over sensitive information (such as PII, commercial confidential information) residing in financial records, health data, legal documents, and other classified documents. This information exists alongside non-sensitive information in data sources (e.g.\ text documents) and requires contextual knowledge to identify and sanitize. In existing literature, various text sanitization methods are discussed that can identify the sensitive content and then redact or anonymize it from source data while keeping the non-sensitive information intact \cite{vasudevan2014review}.

The text sanitization includes a collection of methods used to anonymize/redact sensitive information in data to maintain data privacy. The text sanitization methods are broadly classified into two categories: (i) Reversible redaction and (ii) Irreversible redaction \cite{sarkar2014reversible}. In reversible redaction, the sensitive content is temporarily obscured using masks with the intention of retrieving the content later. This type of redaction is mainly used in communication systems where the sensitive content needs to be obscured only during transit while the sender and receiver need the sensitive information intact for their perusal. Irreversible redaction is prominently used in situations where the sensitive content is required to be completely sanitized and should not be retrievable by any means. Most business transactions require the sensitive content to be completely removed to maintain data privacy during data usage and handling.

The sanitization methods are developed to handle sensitive information that has regular patterns such as PII (telephone numbers, SSN, addresses). Currently these methods use automatic or semi-automatic techniques to identify these contents and redact them. But in most cases sensitive information is defined owing to various business requirements, regulations and mandates which may not have regular patterns like PII. For example, legal clauses, medical diagnoses, and contract terms are given as free text, and identifying sensitive content in them is a challenging task that requires contextual knowledge and expert manual intervention.

Traditionally, irreversible redaction is performed manually using subject matter experts and supporting NLP tools. These methods are labor-intensive and skill-dependent, incurring high costs when the variety and/or volume of data increases \cite{velupillai2009developing}. The quality of redaction directly depends on the efficient identification of sensitive information in the corpus data \cite{lison2021anonymisation}. Multitude methods are available in literature that are used to identify the sensitive data in the given corpus/data, including identification rules and machine learning models. Rule-based sensitive data identification is appropriate for information that has regular patterns such as name, email, date, address, and zip code, but is less efficient for information without regular patterns and involving rich semantics (e.g.\ sentences and phrases in a document).

In typical sentences, each context can be represented using different phrases and forms of speech. A single sensitive (textual) content can be represented in different patterns of phrases, leading to multiple patterns for a single context-sensitive content. This pattern variety increases astoundingly with the variety of sensitive content and its context. Sketch-based approaches that are rule (regular expression) driven require a custom sketch for each distinct pattern of textual data with simple semantics (e.g.\ sets of phrases or sentences in the document). These sketches need to be updated for each distinct representation of sensitive information, leaving such approaches non-scalable for complex data semantics. The limitation of sketch-based methods in handling multiple-pattern content is overcome by machine learning (ML) or deep learning (DL) models that use training data and differential privacy approaches to identify/classify sensitive data in the corpus. DL-based models can learn multiple patterns and can be used to infer any content or similar content that was learned. In differential privacy-based approaches, sensitive content identification involves learning replacement tokens (generalization words) in place of sensitive words/tokens \cite{fung2010privacy}. ML model-based methods are domain-specific and rely on the semantics of the domain of the training data. Hence this approach requires multiple domain-specific ML models to identify sensitive content in rich-semantics corpora, incurring high model management costs \cite{chen2023customized}.

Recent advancements in Large Language Models (LLMs) help overcome the domain-specific knowledge limitations suffered by ML models for identifying sensitive content in large corpora. LLMs are trained over large corpora and hence embed cross-domain knowledge that can be used to identify and redact sensitive content. Traditional redaction tools, which are often rule-based or keyword-driven, fall short when dealing with dynamic and complex contexts, offering limited adaptability and no meaningful explanation for their actions.

This paper presents an intelligent and explainable \textbf{text sanitization solution} that utilizes a smaller fine-tuned language model to perform in-context reasoning over documents. It dynamically generates or selects applicable redaction rules (in natural language) and redacts sensitive information giving explanations in natural language. Unlike black-box machine learning approaches or rigid rule-based systems, this solution combines the interpretability of rule-based logic with the contextual understanding of state-of-the-art natural language models. This hybrid method allows the system to adaptively understand new document domains (not encountered during training), perform redaction, and explain redaction actions in human-readable terms. The solution is evaluated with respect to performance and robustness of redaction. The performance of redaction is evaluated using a manually labeled sample dataset. This model shows high precision in redacting the sensitive content. The robustness of redaction measures the difficulty in reconstructing the redacted document using inference attacks. The evaluation of robustness ensures the coverage of redaction over the entire document or set of documents. This method shows significant evolution in the way of defining sensitive content and redacting it efficiently with less manual intervention.

\section{Preliminaries and Related Work}

Automated text redaction has garnered significant interest due to its applications in data privacy. Especially with the enforcement of regulations like the GDPR and HIPAA, data privacy is mandated to ensure the privacy of the subjects. Traditionally, the process of anonymizing sensitive content is performed manually by subject matter experts (SMEs). Manual redaction depends on SME skill, is time-consuming, and costly as content variety and volume increase.

Early approaches relied heavily on regular expressions and static patterns \cite{neamatullah2008automated}, which proved brittle against diverse document content and linguistic variation. Regular expression-based approaches are suitable for sensitive content with regular formats (such as names, phone numbers, addresses, and SSNs). These approaches cannot be used to redact content that has no regular format and has rich semantics (such as one or more sentences or phrases in a document) \cite{microsoft2023presidio}. For example, in a contract document, the explanation of a sensitive archetype may involve a set of meaningful sentences or contextual phrases.

Sketch-based methods abstract the regular pattern of sensitive content and non-sensitive content in a document as standard sketches. These sketches use identifiers for both sensitive and non-sensitive content. The sensitive content is identified by mapping the sketches onto the document. The sketch-based methods' performance is affected by variations in the sensitive content patterns and their representation patterns in the document. Hence these methods are not suitable for rich-semantics textual documents \cite{ye2020sketch}.

Recent advances leverage NLP, machine learning, particularly deep learning-based named entity recognition (NER), to identify sensitive entities in context \cite{dernoncourt2017neural,liu2021privacy}. Transformer-based models such as BERT have significantly improved NER performance \cite{devlin2019bert}, facilitating more accurate redaction of entities like names, addresses, and medical terms. NER-based approaches can effectively identify information with regular patterns such as names and addresses, but are limited in their ability to identify information whose sensitivity is associated with semantics driven by specific requirements. For example, information related to the dimensions of a land plot is deemed sensitive if it is associated with its location information or any legal agreement or contract \cite{pamarth2024ai}.

Differential privacy-preserving approaches use ML models to predict generalized labels that anonymize/redact the sensitive content in the document. These approaches use the domain-specific knowledge of the ML model to redact the sensitive content. They are not flexible for redacting content involving information from multiple domains, and also when the criteria of sensitive content (rules that infer specific content to be sensitive) are dynamically changed. The former requires ML models trained over multiple domains, while the latter incurs repetitive training for every criterion defining the sensitivity of the content, including data preparation and SME re-labelling \cite{chen2023customized}.

Recent advancements in LLMs with cross-domain knowledge make them suitable for various text re-writing and de-identification tasks. Such LLM knowledge can be used to identify and anonymize sensitive content in the document by instructing the model about what content is deemed sensitive \cite{pilan2025truthful}. This method requires re-tuning of instructions if the criterion of sensitive content is modified. Hence in this article a generalized method is proposed that instruction-tunes the LLM to understand and/or reason about various criteria of content sensitivity that can adapt during inferencing for any criteria of redaction. These criteria are contextualized by the LLM as rules of redaction which are reasoned step-by-step on the input document to identify the sensitive content, later redact it, and mask it with a generalized label (e.g.\ \texttt{[Redacted]}). This solution uses the knowledge and reasoning of the LLM to redact the sensitive content.

Until recent studies, the evaluation of redaction performance was majorly qualitative, relying on SME expertise \cite{pilan2022tab}. With the increase in volume of document transactions, the evaluation uses ML-based metrics such as recall and precision to measure the performance of redaction. The quality of redaction should be balanced between the removal of sensitive content (for privacy) and the preservation of non-sensitive content (for utility) of the document \cite{pilan2025truthful}. The solution also provides a detailed natural language explanation of the redaction decision, which can be used by Responsible AI practices and audit professionals to validate privacy preservation. The explainability of the redaction process involves the LLM generating the explanation (in natural language) of the redaction decision as part of the process itself. The next section discusses the detailed process of data redaction, evaluation of its performance and robustness.

\section{Methodology}

The current methods that use LLMs to identify sensitive content within documents mainly use labeling of the textual span comprising personal information. This labelled dataset is used to train the LLM to identify and redact sensitive content in the document \cite{pilan2022tab,pilan2025truthful}. In this article the proposed method is not limited to handling only personal information but any content that is declared as sensitive by the user.

The method includes three major steps: (i) create synthetic rules of redaction for a given set of domains and/or documents, (ii) instruction-tune a smaller LLM using these synthetic rules and documents, giving the LLM its skill to identify sensitive content and prompt it to redact, and (iii) use the skill-trained LLM to redact sensitive content in given documents during inference.

Firstly, the method comprises a general-purpose LLM that synthetically generates various rules of redaction for a given domain and/or document. The LLM is prompted to create these rules. The rules of redaction represent the criteria that define the sensitivity of specific content in large documents. These rules cover as many possible criteria of sensitivity for any given domain document as possible. The synthetic rules form a critical aspect that helps skill-train the smaller LLM (e.g.\ Phi~3.5), giving it the skill to identify sensitive content in any domain, even those not explicitly learned during the training phase. These rules are in natural language and can be used to instruction-tune the smaller LLM to learn step-by-step and understand the rules of redaction while parsing any given document. Prior to training, the document to be parsed is checked for formatting wellness. The given document needs to be clean in terms of formatting, which enables precise parsing of content.

During inference, the LLM parses the given document based on the rules of redaction supplied by the user. While parsing, the solution identifies the sensitive content in the document by reasoning the rules of redaction over the document content. Each sensitive content (which may be words or phrases) identified by such reasoning is redacted and the redacted span is substituted with generalized tokens (e.g.\ \texttt{[Redacted]}) that indicate the redaction span. The solution redacts only the sensitive content that is relevant (to rules of redaction) while leaving other content intact. The overall solution operates on three functions:

\begin{itemize}
  \item \textbf{Rule Extraction and Redaction:} A fine-tuned small LLM is leveraged to infer context-specific rules for identifying sensitive content and supporting step-by-step in-context reasoned redaction decisions that are grounded in the semantics of the document content.

  \item \textbf{Explainability Layer:} Each redaction is accompanied by a rationale explaining in natural language \textit{why} the information is sensitive and \textit{what rule or guideline} it violates (e.g., ``This is a full Social Security Number, which constitutes PII under GDPR''; ``This content discusses the agreed contractual term, and it is commercially sensitive information'').

  \item \textbf{Human-Auditable Pipeline:} The system produces a redacted document and a trace of rules of redaction. This information supports compliance auditing and reviewer validation.
\end{itemize}

\begin{figure}[t]
  \centering
  \includegraphics[width=\linewidth]{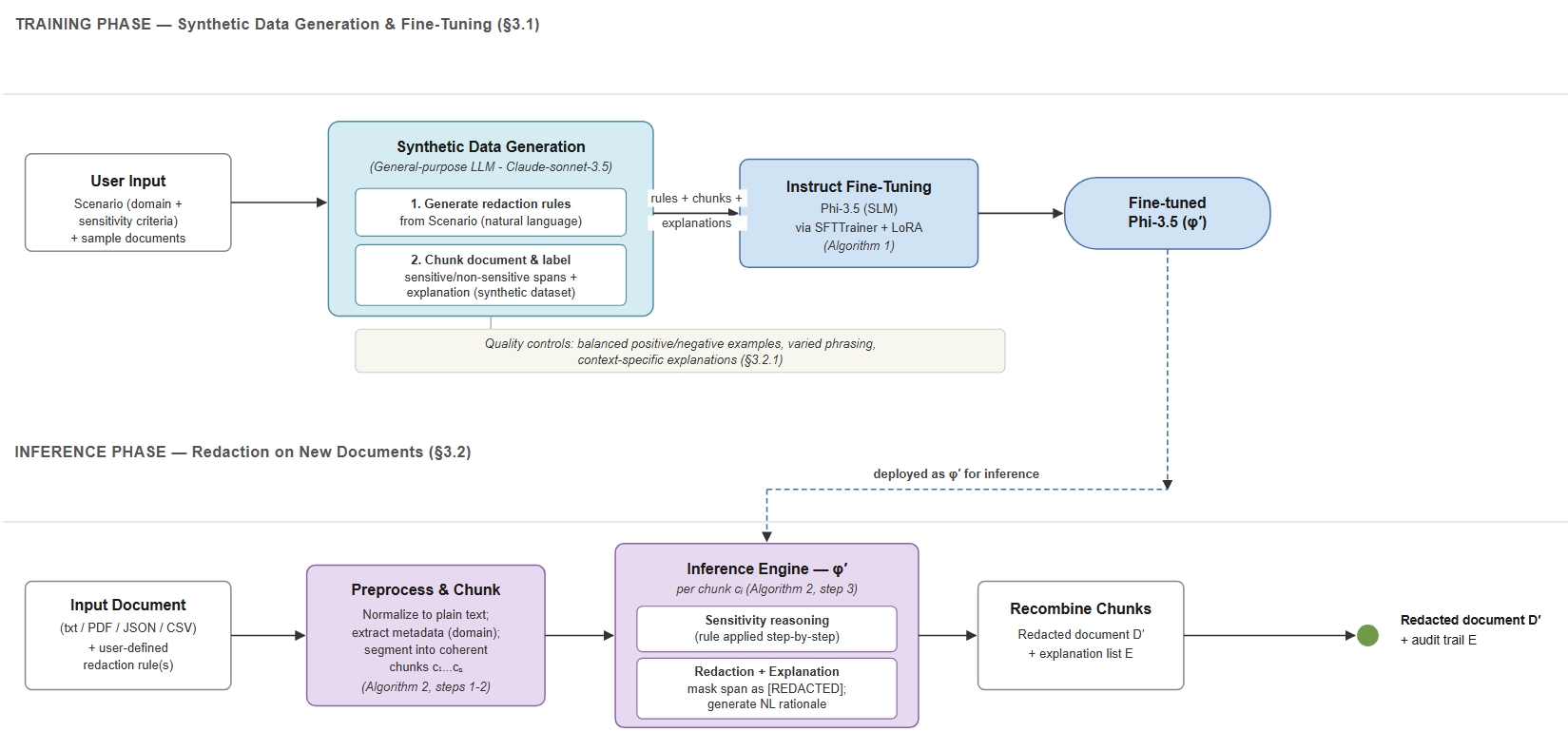}
  \caption{Framework for text sanitization using fine-tuned LLM.}
  \label{fig:framework}
\end{figure}

The objective is to develop a functional pipeline that enables end-users---such as a Chief Risk Officer (CRO)---to define a redaction scheme through a set of rules, all in natural language. The system interprets these rules to identify and redact sensitive information from documents automatically. Designed to be domain-agnostic, the solution can handle a wide range of document types---from financial reports to medical records---thereby supporting regulatory compliance across diverse sectors. Figure~\ref{fig:framework} shows the basic functional blocks used to infer the rules of redaction for a given document and later redact the content.

\subsection{Building Blocks of the Text Sanitization Framework}

The major functional blocks are: Data Source, User Input Interface, Synthetic Data Generation, Fine-tuning, Inference Pipeline, and Evaluation Pipeline.

\textbf{Data Source:} The data source comprises free-text documents. The content of the documents is in various type settings and may include information from more than one domain. The document may be in various formats such as \texttt{.txt} or JSON. For example, a medical insurance claim document comprises medical information and insurance-related content. At the document level it may span different domains including finance, business, medical, and legal. The content comprises domain-specific semantics---the same content across different domains has different semantics and perception of privacy. This perception of privacy and semantics govern the \textit{Scenario}, which represents the domain and criteria of sensitivity of specific content. For example, specific content such as payment, fees, and expenses may be sensitive with respect to business/finance (Scenario~1) whereas they are not as sensitive in legal domains (Scenario~2). This shift in perception of privacy and semantics requires the text sanitization method to be flexible in adapting to various scenarios. The scenario of the documents is realized from the criteria defined by the user from the perception of privacy of specific content, derived or mandated from privacy rules and regulations.

\textbf{User Input Interface:} The user is enabled to supply---through prompts---the documents to be redacted and the Scenario (representing domain, criteria of sensitivity and rules of redaction). The rules of redaction are generated using general-purpose LLMs such as Anthropic Claude for a given scenario and document. The rules of redaction and the document are used to instruction-fine-tune a smaller LM to infer rules of redaction and reason them step-by-step over the input document.

\textbf{Synthetic Data Generation:} The synthetic data generation module is used to generate training datasets from the set of documents and scenarios that represent domain and criteria of sensitivity. Document preprocessing includes chunking the content into fixed-length segments. The chunking size may be uniform and the rules of redaction are applicable to the entire document. This module is powered by general-purpose LLMs (Claude-sonnet), which create the set of rules of redaction and appropriately redact sensitive content (within each chunk of document) with an explanation for each redaction decision. This information is constituted to form the training dataset.

\textbf{Fine-tuning:} The fine-tuning uses the training dataset comprising rules of redaction, sensitive content of documents, and explanations. The smaller language model Phi~3.5---powerful and efficient in reasoning-related tasks---is instruction-fine-tuned to interpret rules of redaction and reason them over the processed content of the document. This fine-tuned model learns the various patterns of sensitive content with respect to rules of redaction.

\textbf{Inference Pipeline:} The inference pipeline constitutes the instruction-fine-tuned Phi~3.5 deployed to infer on the rules of redaction on the given processed input document (chunks). The inference pipeline inputs include the given document, user-defined rules of redaction, and Scenario. The output of the inference pipeline is the redacted document devoid of sensitive content mandated by rules of redaction. Post-processing document chunks after redaction by Phi~3.5 includes merging the chunks into a single document. Such redacted documents are then evaluated for redaction performance.

\textbf{Evaluation of Performance:} The evaluation module measures the correctness of redaction using manually labelled datasets. Correctness represents appropriate redaction exactly in accordance with rules of redaction, and measures the probability of identifying sensitive content in the entire document. The reconstruction metric is used to evaluate the coverage of redaction---representing whether all actual instances of sensitive content within the entire document are redacted. A lower reconstruction metric indicates higher coverage of redaction. The evaluation intends to balance the redaction between sensitive and non-sensitive content (privacy and utility respectively) of the document.

\subsection{Training Phase}

\subsubsection{Synthetic Data Generation}

This phase focuses on constructing a robust pipeline for synthetic data generation to support the fine-tuning of Phi~3.5 (smaller) language model. It employs a carefully engineered prompt template and leverages Claude-sonnet-3.5 to generate high-quality synthetic data spanning a diverse set of domains and use-case scenarios (privacy criteria). This approach ensures comprehensive coverage of relevant linguistic constructs and context-driven sensitive content identification task variations, thereby improving the model's generalization and effectiveness in redaction skills.

For the experimental requirements of this work, the synthetic dataset is curated across the following key domains:

\begin{itemize}
  \item \textbf{Financial Information:} e.g., RFP documents disclosing company performance, revenue figures, or historical financial metrics.
  \item \textbf{Client Engagement Details:} e.g., project-specific information including client names, deliverables, engagement timelines, and performance results.
  \item \textbf{Compliance and Regulatory Information:} e.g., references to regulatory adherence, audit outcomes, certifications, or compliance practices in client-facing documents.
  \item \textbf{Proprietary Methodologies and Intellectual Property:} e.g., descriptions of internal frameworks, methodologies, or tools that may reveal trade secrets or confer competitive advantages.
  \item \textbf{Risk or Issue Management:} e.g., narratives around project risks, mitigations applied, and issues encountered during client engagements.
\end{itemize}

For each domain, representative scenarios were defined and the LLM was prompted to create synthetic documents along with sets of rules of redaction applicable to those documents. The rules of redaction are derived from the inherent knowledge of the LLM. The synthetic dataset comprises document structures with sensitive and non-sensitive content synthesized in a balanced manner. Content with sensitive information and labelled rules of redaction constitute positive examples, while non-sensitive content mapped with rules of redaction are negative examples. Both types of examples are used to fine-tune the smaller language model (SLM) Phi~3.5.

\paragraph{Data Quality and Control Criteria}

To maintain the integrity and utility of the synthetic dataset, several key quality aspects are maintained:

\begin{itemize}
  \item Inclusion of a substantial proportion of negative examples to train the SLM to avoid over-redaction (redacting non-sensitive content).
  \item Ensuring variety and nuance in generated natural language redaction rules, capturing different syntactic and semantic patterns in the content.
  \item In cases where no sensitive information is present, the model should return the content intact without any trace of redaction.
  \item Generated explanations for the reason/decision of redactions must be specific to the document context, avoiding vague or generic justifications.
  \item Only the information explicitly identified as sensitive in the context of the rule should be redacted---ensuring precision and relevance.
\end{itemize}

\subsubsection{Instruct Fine-Tuning}

In this step, Phi~3.5 SLM is fine-tuned over the training dataset. The Phi~3.5 model is known for its efficiency and strong performance on reasoning and alignment tasks despite having fewer parameters compared to large-scale models like GPT-4. Phi~3.5 serves as the foundation model of the redaction system, where it must interpret diverse, user-defined natural language rules and accurately apply them to redact sensitive content from domain-agnostic documents.

While the pretrained Phi~3.5 model demonstrates general language understanding, it lacks the domain sensitivity and instruction adherence necessary for high-stakes redaction tasks. For example, it may fail to recognize contextual identifiers or misinterpret user instructions when dealing with specialized content such as financial disclosures or clinical notes. To address this limitation, \textbf{instruct fine-tuning}---a form of supervised learning---is performed. In this process, the model is trained to follow instruction-style prompts and generate outputs aligned with ground-truth expectations.

The model receives redaction prompts (e.g., ``Remove all personal and financial identifiers'') paired with original documents (processed into chunks) and corresponding human-annotated redacted (ground truth) versions. The model's predictions are continuously compared to the ground-truth redactions, and its weights are updated based on the computed loss. This iterative refinement continues across the dataset, enabling the model to better understand both the explicit and implicit logic behind redaction decisions.

The training is implemented using Hugging Face's \textbf{SFTTrainer} and integrates \textbf{LoRA} (Low-Rank Adaptation) \cite{lison2021anonymisation_models} to fine-tune the model efficiently. LoRA enables adaptation of the model with minimal computational overhead by updating only a small set of additional parameters, keeping the core model mostly intact. This approach retains Phi~3.5's general linguistic capabilities while specializing it for the redaction task.

Post fine-tuning, an improved alignment between the model's output and human expectations is observed, with increased precision and recall in redacted spans, and a better grasp of context-sensitive instructions. This tuning step transforms a general-purpose model into a task-aware system capable of executing nuanced redaction policies. Table~\ref{tab:algo1} shows the algorithm for instruct fine-tuning with LoRA.

\begin{table}[t]
\centering
\caption{Algorithm 1: Instruct Fine-Tuning with LoRA}
\label{tab:algo1}
\begin{tabular}{p{0.95\linewidth}}
\toprule
\textbf{Input:}\\
\quad $\phi$: Pretrained Phi-3.5 model\\
\quad $T$: Training set of pairs $(r, d)$ where $r$ = natural language redaction rule and $d$ = original document\\
\quad $G$: Ground-truth labels for each $(r, d)$ pair where $G = (\text{is\_sensitive}, \text{sensitive\_spans}, \text{explanation}, \text{redacted\_text})$\\[4pt]
\textbf{Output:}\\
\quad $\phi'$: Fine-tuned redaction model\\[4pt]
\textbf{Process:}\\
1.\ Initialize LoRA-adapted model $\phi' \leftarrow \text{apply LoRA to } \phi$\\
2.\ For each $(r, d), G$ in training set $T$:\\
\quad 3a.\ Construct instruction-style prompt $p \leftarrow \text{concatenate}(r, d)$\\
\quad 3b.\ Expected output $y \leftarrow G$\\
\quad 3c.\ Model output $\hat{y} \leftarrow \phi'(p)$\\
\quad 3d.\ Compute loss $\mathcal{L} \leftarrow \text{Loss}(\hat{y}, y)$ where $y$ and $\hat{y}$ include: is\_sensitive (boolean), sensitive\_spans (text or token-level span list), explanation (text), redacted\_text (text)\\
\quad 3e.\ Backpropagate $\mathcal{L}$ and update only LoRA parameters in $\phi'$\\
4.\ Repeat Step 2 for multiple epochs until model performance stabilizes\\[4pt]
\textbf{Return:} $\phi'$ as the fine-tuned, instruction-following redaction model\\
\bottomrule
\end{tabular}
\end{table}

\subsection{Inference Phase}

The inference framework operates in five main stages: \textbf{preprocessing, chunking, sensitivity identification, redaction engine,} and \textbf{explanation generation}, each enhanced by the reasoning capability of our fine-tuned Phi~3.5 SLM.

\textbf{Preprocessing:} Input documents (plain text, PDF, or structured formats like JSON/CSV) are converted into a normalized text representation. Metadata such as document type (legal, medical, financial) is extracted when available to inform the LLM's in-context understanding.

\textbf{Chunking:} To ensure robustness across documents of varying lengths, the pipeline incorporates a chunking mechanism that segments the content into smaller, semantically coherent units. This approach preserves contextual integrity within each chunk while enabling efficient processing. In the final stage, post-redaction, the redacted chunks are seamlessly recombined to reconstruct the complete redacted document, maintaining both continuity and fidelity to the original structure.

\textbf{Sensitivity Identification:} Leveraging the natural language rules provided by the user, the instruction-fine-tuned Phi~3.5 model performs contextual analysis on each document chunk to accurately identify content that qualifies as sensitive. The model interprets the intent behind the rules and applies them semantically, enabling precise and context-aware sensitive content detection.

\textbf{Redaction Engine:} Each identified sensitive content is redacted and the redacted span is masked using token-level replacements or annotations (e.g., \texttt{[REDACTED]}). The system allows configurable granularity (partial vs.\ full redaction) and redaction types (masking, deletion, token replacement).

\textbf{Explanation Generation:} For every redacted item, the SLM generates a short explanation detailing: what the data is (e.g., ``Phone number''), why it is sensitive (e.g., ``can be used to identify or contact an individual''), and which rule or policy it violated (e.g., ``violates GDPR Article 4''). These explanations are inserted as inline annotations or appended to a redaction summary report.

Table~\ref{tab:algo2} shows the algorithm for the inference framework.

\begin{table}[t]
\centering
\caption{Algorithm 2: Inference Framework with Phi~3.5}
\label{tab:algo2}
\begin{tabular}{p{0.95\linewidth}}
\toprule
\textbf{Input:}\\
\quad $\phi'$: Fine-tuned Phi-3.5 redaction model\\
\quad $r$: Natural language redaction rule\\
\quad $D$: Input document (plain text or structured format)\\[4pt]
\textbf{Output:}\\
\quad $D'$: Redacted document\\
\quad $E$: Explanations for each redacted segment\\[4pt]
\textbf{Process:}\\
1.\ Preprocessing:\\
\quad a.\ Normalize $D$ to plain text format\\
\quad b.\ Extract metadata if available (e.g., domain)\\
2.\ Chunking:\\
\quad a.\ Segment $D$ into coherent chunks $C = \{c_1, c_2, \ldots, c_n\}$\\
3.\ For each chunk $c_i \in C$:\\
\quad a.\ Construct prompt $p \leftarrow \text{concatenate}(r, c_i)$\\
\quad b.\ Output $\leftarrow \phi'(p)$ where Output $=$ \{is\_sensitive: bool, sensitive\_spans: list, explanation: list, redacted\_text: transformed $c_i$\}\\
\quad c.\ If Output.is\_sensitive: Append redacted\_text to $D'$; store explanation in $E$\\
\quad\quad Else: Append original $c_i$ to $D'$\\
4.\ Recombine redacted chunks to reconstruct $D'$\\[4pt]
\textbf{Return:} $D'$ (final redacted document), $E$ (explanation list)\\
\bottomrule
\end{tabular}
\end{table}

\section{Experiments and Results}

The SLM model can identify the sensitive content span in the given document and redact the content by labelling \texttt{[Redacted]}. The SLM generates the entire redacted document by post-processing (merging) the document chunks created during the inferencing pipeline. The redacted document comprises explanations in natural language mentioning the reason for redaction of each specific content in the given document. The evaluation of redaction performance is required to assess the quality of redaction. This quality evaluation includes ensuring the balance maintained between redacting the sensitive content and leaving intact the non-sensitive content of the document. The former preserves privacy, and the latter ensures the utility of the document \cite{pilan2025truthful}. In this article, both qualitative and quantitative evaluations are performed, allowing users to understand the quality of redaction both in terms of privacy and utility.

\subsection{Qualitative Evaluation}

For evaluating the effectiveness of the redaction solution, both qualitative and quantitative assessments of the model's output are conducted. The qualitative evaluation involves subject matter experts manually examining the rules of redaction and the redacted document against the ground truth (original document).

Due to the absence of any suitable open-source benchmark datasets \cite{pilan2022tab} for this type of data redaction, a custom ``validation dataset'' is created in this article. This includes manually human-redacted sets of documents derived from diverse domains. Each data point in the dataset includes the following key components:

\begin{itemize}
  \item \textbf{Domain} -- The category to which the document belongs (e.g., healthcare, finance).
  \item \textbf{Rule} -- The natural language instruction provided by the user to guide redaction.
  \item \textbf{Original Document} -- The unaltered source document prior to redaction.
  \item \textbf{Manually Redacted Content} -- The ground truth version, created through human redaction (annotation).
  \item \textbf{Model-Redacted Content} -- The output produced by the fine-tuned redaction model.
  \item \textbf{Explanation} -- A detailed explanation referring to the reason for anonymizing or redacting specific fields in the content or document.
\end{itemize}

The examination by subject matter experts reveals that the redaction is performed only on sensitive content while the non-sensitive content is left intact in the redacted document. The span of sensitive content and the span of redacted content are the same in at most 99\% of the sample data points evaluated. The qualitative evaluation results need to be validated using a quantitative evaluation process, whereby user and audit practices assure the quality of performance.

\subsection{Quantitative Evaluation}

To evaluate the performance of solutions on the validation dataset, two primary metrics are used: \textbf{Reconstruction Accuracy} and \textbf{Redaction Recall}, which respectively measure the resilience of SLM Phi~3.5 against inference attacks and the model's ability to adhere to redaction rules.

\subsubsection{Reconstruction Accuracy}

To evaluate the privacy strength of the redaction solution, we quantify the amount and/or quality of sensitive information an adversarial large language model (LLM) can recover from the redacted documents. A \textbf{reconstruction-based metric} inspired by prior model-inversion attacks is used for quantifying redaction quality \cite{zhang2022text}. This metric measures the \textbf{Coverage} of redaction---meaning whether the redaction is complete and does not omit any part of sensitive content in the entire redacted document. Lopresti and Spitz demonstrated that even redacted documents can unintentionally leak sensitive information through structural or contextual constructs such as predictable formatting, surrounding text patterns, or incomplete removal, highlighting that redaction alone does not guarantee privacy, even in purely textual content \cite{lopresti2004quantifying}.

In this article, a strong adversary attack is simulated using Claude-sonnet-3.5. The redacted document is input to Claude-sonnet-3.5 and it is prompted to reconstruct the redacted span. The LLM's output is then compared to the original (ground-truth) content. \textbf{Reconstruction Accuracy} is defined as the fraction of originally redacted tokens that the LLM correctly recovers. If $N$ is the total number of sensitive tokens removed by redaction and $M$ is the number of those tokens that the LLM's reconstruction matches exactly, then:

\begin{equation}
  \text{Reconstruction Accuracy} = \frac{M}{N}
\end{equation}

A \textbf{higher} reconstruction accuracy (maximum value of 1) means the adversary recovered all sensitive data (poor or no privacy preservation, potentially more information leakage), whereas a \textbf{lower} accuracy (less than 0.1) indicates stronger redaction. This metric is analogous to the ``\textit{Recovery Rate}'' metric used by Zhang et al.\ in their text-reconstruction attack \cite{zhang2022text}, which measures the percentage of private tokens recovered by the attacker. Accordingly, a well-performing redaction model is expected to yield a recovery rate near zero for any inference adversarial LLM attack.

\begin{figure}[t]
  \centering
  \includegraphics[width=\linewidth]{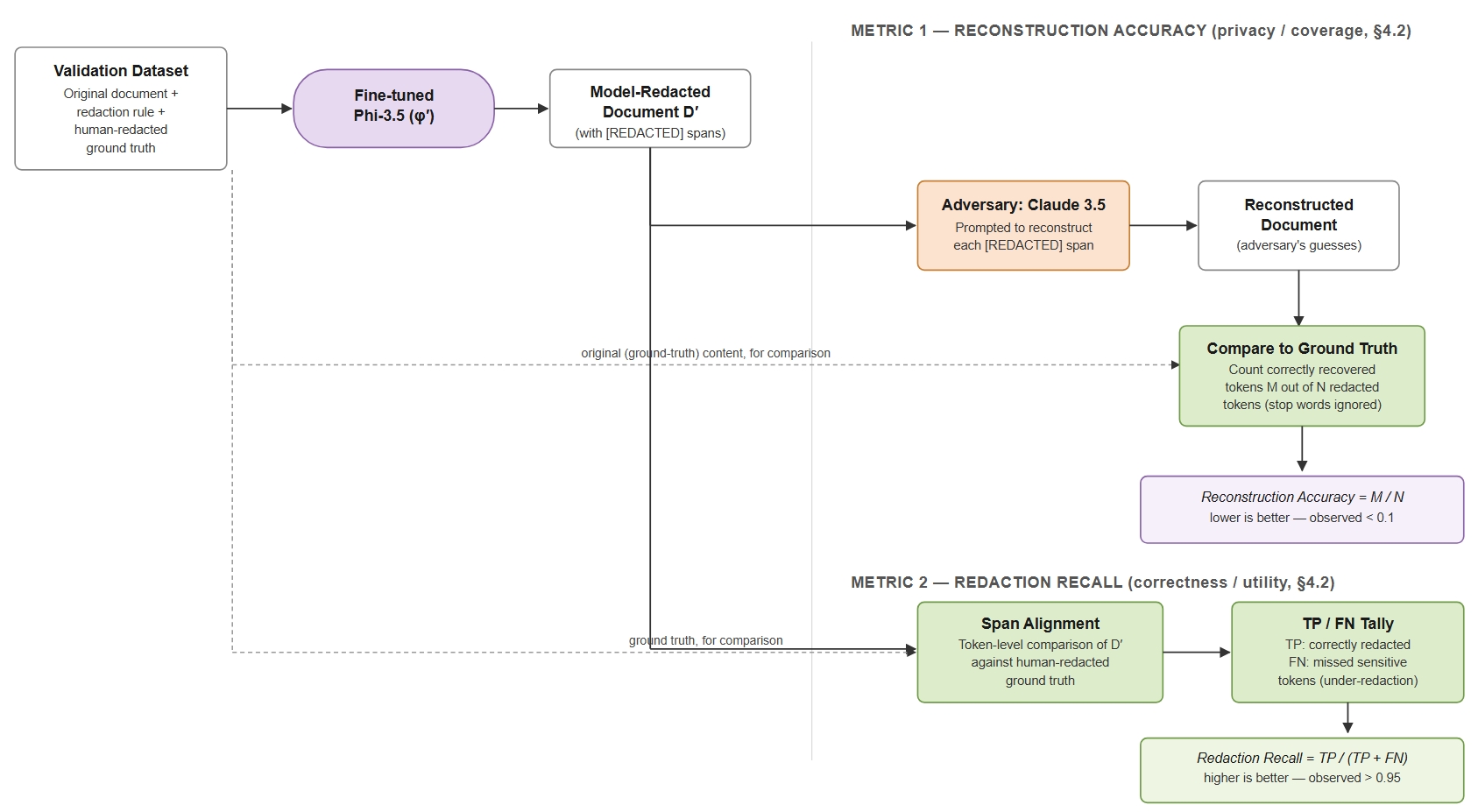}
  \caption{Evaluation framework for the redacted content.}
  \label{fig:evaluation}
\end{figure}

The reconstruction accuracy is evaluated through the following procedure. Figure~\ref{fig:evaluation} shows the evaluation framework for assessing reconstruction accuracy of the redacted content:

\begin{itemize}
  \item \textbf{Adversarial Reconstruction:} The redacted content (by SLM Phi~3.5) in the validation dataset is input to Claude-sonnet-3.5 along with a reconstruction prompt. Claude acts as an adversary that sees the context surrounding the redacted tokens and tries to infer the sensitive content.
  \item \textbf{Comparing Output to Ground Truth:} Claude's reconstructed content is compared with the original (pre-redaction) document content. The text is aligned and the number of originally redacted tokens (e.g.\ named entities, numbers, confidential phrases) correctly recovered is counted. Trivial tokens such as common stop words or punctuation are ignored, focusing only on actual sensitive terms.
  \item \textbf{Metric Computation:} Reconstruction Accuracy $= \frac{\text{Correctly recovered tokens}}{\text{Total redacted tokens}}$. This metric is aggregated across all data points in the sample validation dataset (comprising various domain documents) to obtain an overall score. A lower reconstruction accuracy indicates better privacy, meaning the adversarial LLM failed to guess the sensitive information.
\end{itemize}

The reconstruction accuracy is less than 0.1 for the redacted document, ensuring high resilience of the SLM towards Claude LLM-based inference attacks. This shows that the instruction-fine-tuned Phi~3.5 model has a highly scalable redaction skill while performing equally well in cross-domain document instances.

\subsubsection{Redaction Recall}

The Reconstruction Accuracy is used to evaluate the SLM's robustness to adversarial attacks. The \textbf{Redaction Recall} metric is used to assess how thoroughly the model identifies and redacts sensitive information in alignment with human judgment. This metric captures the \textbf{correctness} of the model's redaction capability and is essential for determining whether the model is omitting any content that should have been removed for privacy.

Redaction Recall is defined as the proportion of all ground-truth redacted tokens---such as named entities, numerical identifiers, and confidential phrases/sentences/paragraphs---that the model successfully redacts. In other words, it quantifies: ``\textit{Of all the content that should have been redacted, how much did the model actually redact?}''

Consider:
\begin{itemize}
  \item \textbf{TP (True Positives):} The number of tokens correctly redacted by the model, as verified against the human-annotated redacted content.
  \item \textbf{FN (False Negatives):} The number of sensitive tokens present in the ground-truth redaction that the model failed to redact.
\end{itemize}

The Redaction Recall is then computed as:
\begin{equation}
  \text{Redaction Recall} = \frac{\text{TP}}{\text{TP} + \text{FN}}
\end{equation}

A high Redaction Recall indicates that the model is effectively capturing the full scope of sensitive content that needs to be redacted. Conversely, a low value reflects under-redaction and raises concerns about residual privacy risk due to missed entities. To compute Redaction Recall, the following procedure is used:

\begin{itemize}
  \item \textbf{Span Alignment:} For each document, the model-redacted output is compared with the human-redacted ground truth at the token level. Only spans of tokens explicitly marked for redaction in human annotation are considered.
  \item \textbf{Token Comparison:} The number of tokens that overlap between the fine-tuned SLM output and the ground truth redaction (TP) is counted, along with the number of tokens that were missed (FN).
  \item \textbf{Recall Aggregation:} Redaction Recall is computed for the entire document and aggregated across the entire dataset (set of documents) to measure overall performance.
\end{itemize}

These metrics are particularly important in real-world settings, where missing even a single sensitive token can lead to a privacy breach. The fine-tuned SLM can redact sensitive content in over 95\% of the sensitive instances in the input document. In this article, various measures are used to ensure the correctness and coverage of redaction, ensuring the data privacy preservation productivity of the SLM, especially in large-volume documents.

\section{Summary}

The solution offers state-of-the-art capability to handle (identify and redact) non-regularly structured information (e.g.\ legal clauses, medical information) in contrast to current text sanitization solutions that can handle only regularly structured information (e.g.\ PII). The critical contribution of this solution includes enabling the user to custom-define any sensitive content---which may be a legal clause, medical therapy, or business contract deal---and the LLM engine automatically generates rules of redaction to redact the sensitive content from the document. The SLM engine reasons the rules of redaction step-by-step on any given document to identify and redact the sensitive content. The SLM engine is trained with synthetically generated sample data from various domains such as Legal, Business, and Medical. The usage of synthetic datasets to train the SLM engine mitigates any privacy attack (inference attacks) on the SLM engine, thereby maintaining data privacy during both training and inference stages.

The SLM engine shows a higher degree of redaction performance with well-formatted documents. In cases of ill-formatted documents (with large blanks and spaces), the solution comprises a data preprocessing pipeline that makes the data amenable for inferencing. The solution uses two dedicated metrics to evaluate the correctness and coverage of the redaction. A customized manually redacted dataset is used to validate the testing performance of redaction. The reconstruction metric gives the coverage performance of the SLM engine throughout the entire document. The reconstruction metric uses a reconstruction attack on the redacted document, evaluating whether any part of the redacted content can be reconstructed. The solution shows high reconstruction errors, which indicate good redaction coverage.

The SLM engine can also be used to redact other domain documents (those not used in training) with minimal few-shot learning and/or fine-tuning. The solution is versatile for redacting any free-text sensitive content that can be custom-defined by the user on-the-fly, and the SLM engine can automatically redact it. The automated data redaction improves the data privacy productivity of sanitizing large documents with less manual intervention, high accuracy, and high utility.

\section*{Acknowledgments}

\bibliographystyle{unsrt}
\bibliography{references}

\end{document}